\documentclass[preprint,12pt]{elsarticle}

\usepackage{amssymb}
\usepackage{amsmath}

\usepackage{booktabs}
\usepackage{subfigure}
\usepackage{pgfplots}
\pgfplotsset{compat=1.16}

\usepackage{amssymb}
\usepackage{multirow}
\usepackage{varwidth}
\usepackage{tikz}

\usepackage{verbatim}

\usepackage{makecell}

\journal{Nuclear Physics B}

\begin{document}

\begin{frontmatter}



\title{Fact in Fragments: Deconstructing Complex Claims via LLM-based Atomic Fact Extraction and Verification}


\author[label1]{Liwen Zheng}
\ead{zhenglw@bupt.edu.cn}

\author[label1]{Chaozhuo Li}
\ead{lichaozhuo@bupt.edu.cn}

\author[label2]{Zheng Liu}
\ead{zhengliu1026@gmail.com}

\author[label3]{Feiran Huang}
\ead{huangfr@jnu.edu.cn}

\author[label1]{Haoran Jia}
\ead{jiahaoran@bupt.edu.cn}

\author[label4]{Zaisheng Ye}
\ead{yzs1986@fjmu.edu.cn}

\author[label1]{Xi Zhang}
\ead{zhangx@bupt.edu.cn}

\affiliation[label1]{organization={Key Laboratory of Trustworthy Distributed Computing and Service (MoE)}, 
addressline={\\Beijing University of Posts and Telecommunications},
city={Beijing},
postcode={100876}, 
country={China}}

\affiliation[label2]{organization={Beijing Academy of Artificial Intelligence}, 
city={Beijing},
country={China}}

\affiliation[label3]{organization={Jinan University}, 
city={Guangzhou},
postcode={510632}, 
country={China}}

\affiliation[label4]{organization={Department of Gastric Surgery, Clinical Oncology School of Fujian Medical University, Fujian Cancer Hospital(Fujian Branch of Fudan University Shanghai Cancer Center)}, 
city={Fuzhou},
postcode={350014}, 
country={China}}

\begin{abstract}
Fact verification plays a vital role in combating misinformation by assessing the veracity of claims through evidence retrieval and reasoning. 
However, traditional methods struggle with complex claims requiring multi-hop reasoning over fragmented evidence, as they often rely on static decomposition strategies and surface-level semantic retrieval, which fail to capture the nuanced structure and intent of the claim. 
This results in accumulated reasoning errors, noisy evidence contamination, and limited adaptability to diverse claims, ultimately undermining verification accuracy in complex scenarios.
To address this, we propose \textbf{A}tomic \textbf{F}act \textbf{E}xtraction and \textbf{V}erification (AFEV), a novel framework that iteratively decomposes complex claims into atomic facts, enabling fine-grained retrieval and adaptive reasoning. AFEV dynamically refines claim understanding and reduces error propagation through iterative fact extraction, reranks evidence to filter noise, and leverages context-specific demonstrations to guide the reasoning process. Extensive experiments on five benchmark datasets demonstrate that AFEV achieves state-of-the-art performance in both accuracy and interpretability.
\end{abstract}









\end{frontmatter}




\section{Introduction}
Fact verification is a critical task aimed at assessing the veracity of claims and typically involves two fundamental stages: evidence retrieval and claim verification. During the evidence retrieval phase, relevant pieces of information are identified from large-scale corpora. 
In the subsequent verification phase, the retrieved evidence is synthesized and analyzed to determine the veracity of the claim~\cite{FEVER}. 
Fact verification serves as an essential pillar in the modern information ecosystem, mitigating the spread of misinformation, fostering the integrity of knowledge systems, and supporting informed decision-making in economic and political domains.

Despite significant progress in fact verification, existing approaches primarily focus on verifying relatively simple claims, in which the authenticity can be determined through a single piece of evidence~\cite{2022-survey}. 
However, real-world claims are often more complex, requiring reasoning over multiple interrelated pieces of evidence to reach a reliable conclusion~\cite{PolitiHop, RAV}. 
Multi-hop verification methods attempt to address this challenge by aggregating and reasoning over multiple evidence snippets, demonstrating promising performance in handling complex claims~\cite{Baleen, MRR-FV}. 
Nevertheless, the multi-hop evidence retrieval process frequently introduces a large number of loosely connected sentences, leading to a fragmented context where critical clues may be obscured. 
Conventional reasoning models often struggle to directly address complex verification tasks, as they exhibit limitations in modeling long-range dependencies and performing multi-step inference. 
This necessitates a strong reasoning capability to synthesize dispersed information effectively — a strength that large language models (LLMs) are particularly well-equipped to provide.
The emergence of large language models presents a transformative opportunity for fact verification, particularly in verifying complex claims that demand sophisticated multi-hop reasoning~\cite{Loki}. 
LLMs possess exceptional comprehension abilities and extensive world knowledge, enabling them to extract fine-grained relationships among entities and infer implicit connections across multiple pieces of evidence, which enhances their potential to overcome the limitations of conventional methods.

\begin{figure*}[t]
\centering
\includegraphics[width=1.0
\linewidth]{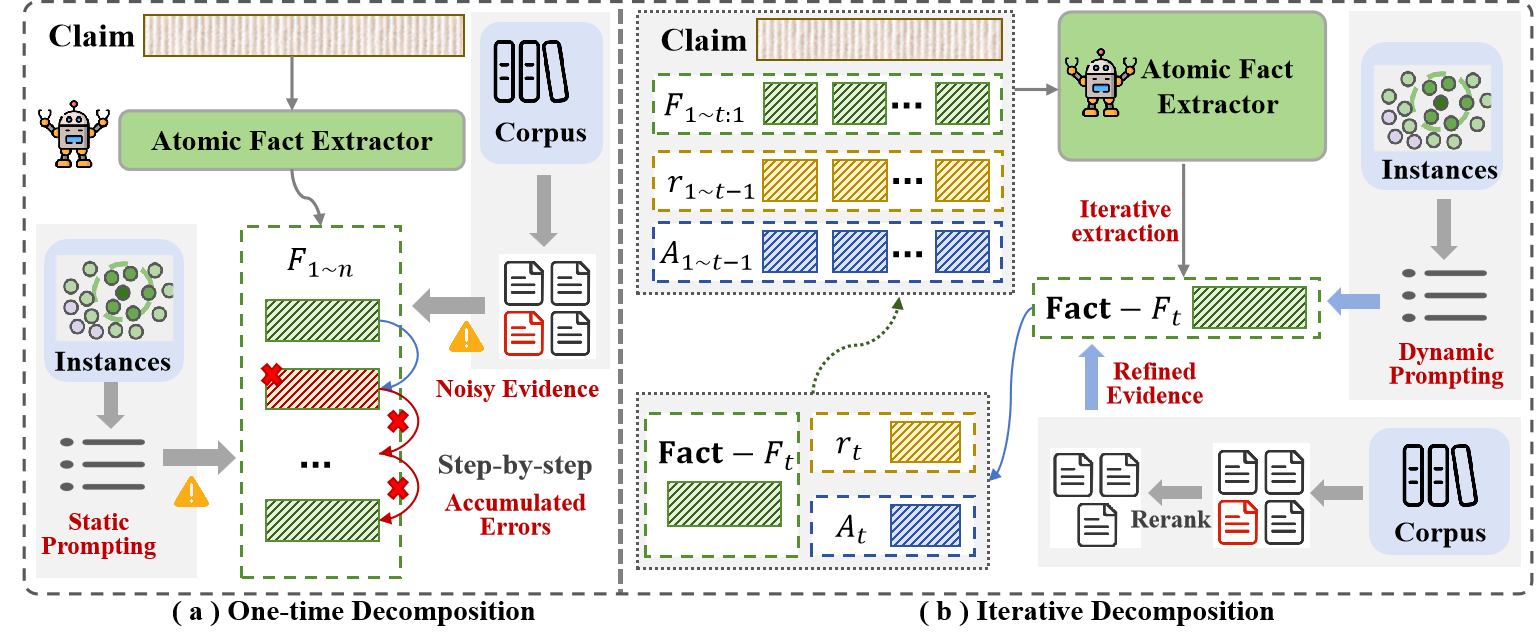}
\caption{Comparison of different claim decomposition strategies.}
\label{intro}
\end{figure*}

Despite the impressive generative capabilities of LLMs, directly applying them to fact verification presents numerous challenges.
A significant limitation stems from inherent issues such as training data biases, parametric overfitting, and knowledge gaps, which collectively contribute to the phenomenon of ``hallucination" –-instances where models generate plausible but factually inaccurate or ungrounded content~\cite{Self-RAG}. 
Therefore, direct reliance on LLMs for factual verification is inherently imprudent and risky.
Some current LLM-based fact verification methods adopt an Iterative Retrieval and Verification strategy. By incorporating external knowledge into the model's context and leveraging feedback mechanisms to validate the factuality, these methods achieve improved accuracy in certain scenarios~\cite{RAC}. Nevertheless, they still exhibit significant limitations when tackling complex fact verification problems that require deeper reasoning and logical integration~\cite{COV}.
To overcome these challenges, an emerging line of research focuses on atomic fact extraction from complex claims. 
As depicted in Figure~\ref{intro}(a), this approach simplifies the problem by breaking down intricate statements into simpler atomic facts, facilitating targeted retrieval and verification of individual components. This decomposition not only enhances the model's ability to handle nuanced logical relationships but also reduces the risk of cascading errors during reasoning.

While existing methods have shown promise in mitigating hallucination risks in large language models, they remain weakly coupled with strong claim understanding. 
Most approaches emphasize acquiring external knowledge, while overlooking the need to improve comprehension of the original claim, which limits their performance on complex cases~\cite{COV}. 
These limitations manifest in the following ways:
1) Accumulated Errors from Static Decomposition Strategies:
Although decomposing complex claims into atomic facts can simplify verification, conventional static strategies—whether rule-based or zero-shot LLM-driven—fail to adapt dynamically to the semantic granularity and contextual dependencies inherent in claims. These approaches tend to prioritize syntactic fragmentation over a contextual understanding of claim intent and structure. Furthermore, the absence of explicit supervision during decomposition amplifies error propagation, especially in multi-hop reasoning scenarios.
2) Noisy Evidence Contamination in Surface-Level Semantic Alignment: Semantic similarity-based fact verification frameworks are inherently vulnerable to context-agnostic retrieval, where surface-level alignment overrides deeper claim-evidence compatibility. This often leads to the inclusion of extraneous evidence fragments and, in multi-hop reasoning, overemphasis on isolated subconcepts while neglecting causal dependencies embedded in the original claim. 
3) Constraints of Static Prompting on Claim Understanding and Reasoning:
Most existing LLM-based verification methods rely on static, complex prompts to guide reasoning. However, verifying diverse claim types requires varied reasoning strategies and deeper contextual understanding—capabilities that static prompts often fail to support. This rigidity limits the model’s capacity to grasp claim intent and structure, leading to shallow or misaligned reasoning. Consequently, the model struggles to generalize across verification scenarios, hindering overall performance.

To address the aforementioned challenges, this paper proposes an \textbf{A}tomic \textbf{F}act \textbf{E}xtraction and \textbf{V}erification (AFEV) framework for reliable fact verification, which progressively decomposes complex claims into simpler atomic facts. 
As illustrated in Figure~\ref{intro}(b), the LLM-enabled extractor transforms intricate reasoning tasks into manageable subtasks, allowing for more precise, targeted verification of individual components. 
To mitigate the error accumulation caused by static decomposition strategies, the framework employs an iterative atomic fact extraction strategy. 
This strategy leverages the understanding of previously verified atomic facts to guide the subsequent decomposition process, ensuring that claims are decomposed more coherently with each iteration.
By integrating fine-grained external evidence and dynamic demonstrations, the proposed approach ensures accurate validation of atomic facts while maintaining both interpretability and reliability throughout the verification pipeline. 
Specifically, the pretrained evidence re-ranker is utilized to filter out irrelevant or noisy information for each atomic fact, prioritizing highly relevant evidence and reducing the impact of distractive data on the verification results.
In addition, instead of relying on static prompting methods, our framework dynamically retrieves relevant instances specific to each atomic fact, tailoring the reasoning process to the unique characteristics of the fact under verification. 
During the atomic fact verification phase, the reasoner generates both a factuality label for each atomic fact and a corresponding rationale, guided by dynamic demonstrations and reranked evidence.
This process leverages LLMs’ capacity for adaptive reasoning and interpretable decision-making.
Finally, the reasoner synthesizes the verified atomic facts to form a comprehensive judgment of overall factuality.
Experimental results on several benchmark datasets validate the superiority of this method.

The main contributions of this paper include:

\begin{itemize}

    \item This paper tackles the challenge of understanding and verifying complex claims by introducing a novel full-cycle fact verification framework. It iteratively decomposes and refines intricate claims into atomic facts, thereby strengthening claim comprehension and markedly improving both accuracy and interpretability.
    \item The proposed AFEV framework comprises three key modules: the Dynamic Atomic Fact Extraction module, which is built upon LLMs and iteratively decomposes complex claims into atomic facts, thereby reducing reasoning complexity and mitigating error propagation; 
    the Refined Evidence Retrieval module, which retrieves fine-grained evidence and dynamically samples demonstrations to guide more reliable reasoning; 
    and the Adaptive Atomic Fact Verification module, which harnesses the advanced reasoning capabilities of LLMs to generate verification results and rationales for each atomic fact, further enhancing subsequent decomposition and verification processes.
    \item Extensive experiments on five fact verification datasets demonstrate the superior performance and effectiveness of AFEV in handling complex claims.
\end{itemize}

\section{Related Work}

Recent research on multi-hop fact verification employs iterative evidence retrieval and reasoning mechanisms to model inference paths and capture critical clues for accurate claim verification~\cite{RACE}.
HESM~\cite{HESM} employs a hierarchical multi-hop evidence retrieval mechanism; however, its heavy reliance on entity links constrains its applicability across different scenarios.
MDR~\cite{MDR} and Baleen~\cite{Baleen} leverage dense retrieval to link claims with previously retrieved evidence, enhancing the original query. RECOMP~\cite{RECOMP} enhances retrieval by explicitly capturing semantic-level interactions across evidence pieces.
During the claim verification stage, existing methods construct multi-hop evidence graphs and perform graph reasoning to extract key clues for assessing claim veracity~\cite{Causal-Walk}. Additionally, some approaches introduce learnable perturbation mechanisms to remove redundant information, enabling more fine-grained reasoning~\cite{SaGP}.

Existing multi-hop fact verification methods are somewhat effective in handling complex claims. 
However, their supervised training paradigm relies heavily on large-scale annotated data, limiting both scalability and cross-domain adaptability.
In contrast, Large Language Models (LLMs) offer a paradigm shift for claim verification by leveraging deep language understanding, multimodal fusion, and logical reasoning. 
Moreover, the Retrieval Augmented Generation (RAG) framework, which dynamically integrates external evidence with internal knowledge, has been shown to mitigate hallucination issues in LLMs, thereby enhancing the accuracy and interpretability of fact-checking~\cite{FOLK,QACHECK}.
Self-RAG~\cite{Self-RAG} introduces a confidence-based dynamic evidence retrieval decision mechanism, while FFRR~\cite{FFRR} innovatively constructs dual-dimensional feedback signals, including document-level semantic integrity and query-level relevance, to iteratively optimize the retrieval model. Meanwhile, CRAG~\cite{CRAG} incorporates a lightweight retrieval quality assessment module that dynamically triggers completion or re-retrieval operations based on probabilistic thresholds. Furthermore, some studies enhance fact-checking by constructing adversarial argument spaces, strengthening model robustness in reasoning under conflicting evidence and advancing the paradigm toward dialectical reasoning~\cite{RAFV, Competing}.

Although the aforementioned methods refine verification results through intrinsic feedback mechanisms, they still face challenges in handling complex factuality tasks. To address this issue, recent studies have focused on dynamic iterative retrieval-verification frameworks, which integrate external knowledge into the model’s context and leverage feedback mechanisms for factuality validation~\cite{COV, Generate}. RAC~\cite{RAC} decomposes complex claims into independently verifiable atomic fact units and employs reinforcement learning to optimize retrieval queries and weight evidence credibility, enabling progressive factual correction. FIRE~\cite{FIRE} introduces an iterative retrieval and verification mechanism, incorporating an Adaptive Retrieval Trigger that dynamically generates multi-granularity retrieval instructions based on evidence completeness metrics. 
Notably, the effectiveness of these approaches hinges on the precision of claim decomposition. Consequently, ongoing research explores methods for breaking down complex claims into more verifiable atomic facts~\cite{Molecular-Facts}.
Nevertheless, existing static claim decomposition strategies struggle to maintain information completeness and coherence, leading to broken reasoning chains and exacerbating redundancy, which in turn increases the risk of error accumulation during the inference process.

\section{Problem Definition}
The proposed framework for fact verification employs an iterative atomic fact extraction process, which decomposes a complex claim \( C \) into a set of atomic facts \( F = \{ F_1, F_2, \dots, F_T \} \). 
For each atomic fact \( F_i \), relevant evidence \( E_i = \{ e_1^i, e_2^i, \dots, e_k^i \} \) is retrieved from an external corpus \( D \), ensuring that the verification process is grounded in factual information. 
To enhance the reasoning module's precision, dynamic demonstrations \( A_i = \{ a_1^i, a_2^i, \dots, a_d^i \} \) are constructed by retrieving highly relevant claims associated with \( F_i \). 
Each demonstration \( a_i \) consists of a claim, its ground-truth label, corresponding gold evidence, and a rationale, providing the reasoning module with contextual guidance and improving its ability to handle nuanced verification tasks.

The verification process for each atomic fact \( F_i \) is formally defined as:
\begin{equation}
y_i, r_i = \text{Reasoner}(F_i, E_i, A_i),
\end{equation}
where \( y_i \) represents the verification result and \( r_i \) denotes the rationale generated by the reasoning module. 
Finally, the verification results and rationales for all atomic facts are aggregated to yield the overall verification label \( y^* \) for the original claim \( C \).

\section{Methodology}
\begin{figure*}[t]
\centering
\includegraphics[width=1.0
\linewidth]{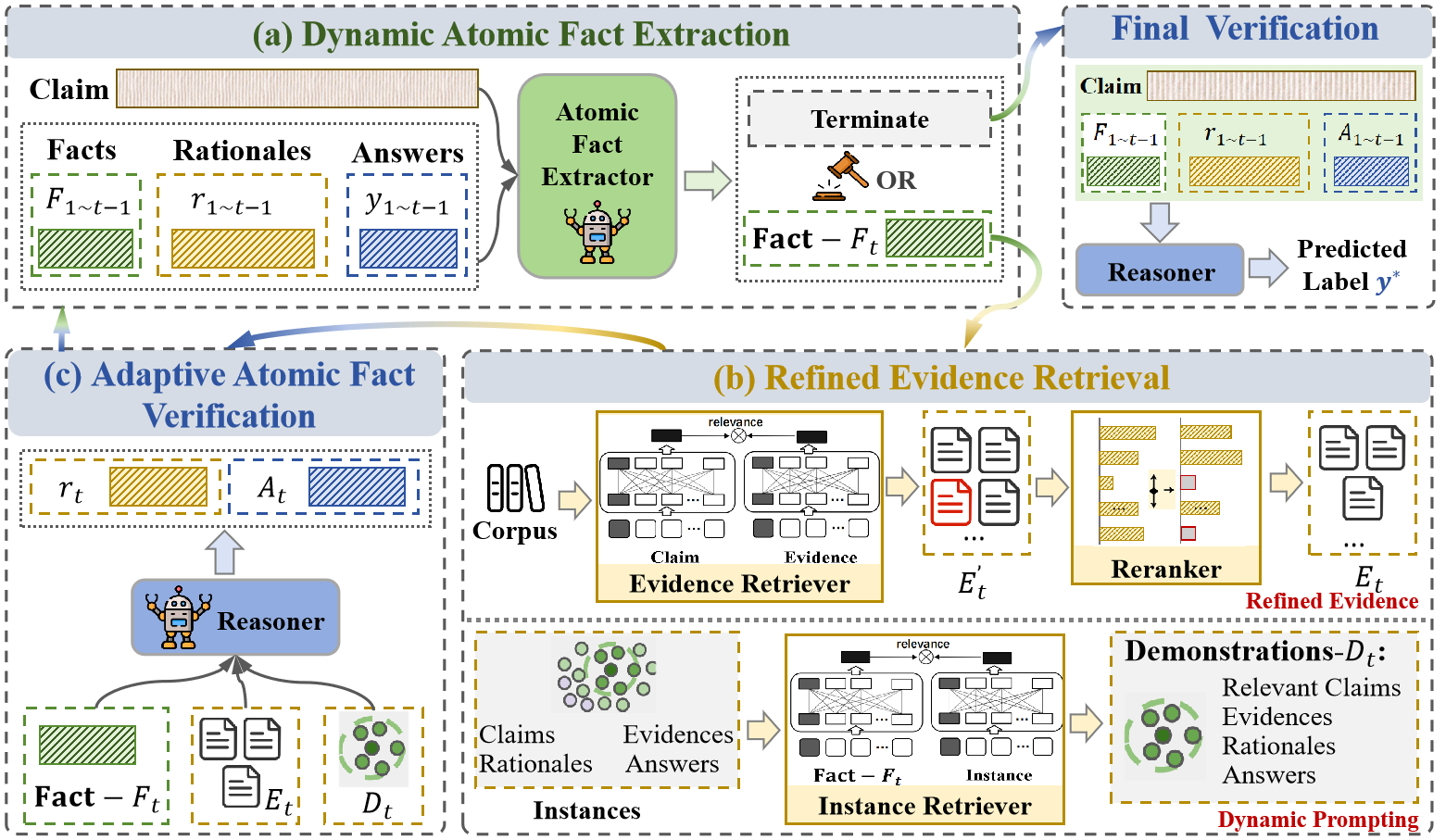}
\caption{Framework of the proposed AFEV model.}
\label{fig:model}
\end{figure*}

As illustrated in Figure~\ref{fig:model}, the proposed AFEV model performs fact verification through a three-stage framework: 1) the Dynamic Atomic Fact Extraction module iteratively decomposes the original claim into atomic facts, simplifying complex reasoning into manageable sub-problems; 2) the Refined Evidence Retrieval module utilizes a pretrained evidence reranker to eliminate irrelevant information and retrieve highly relevant instances as dynamic reasoning demonstrations; and 3) the Adaptive Atomic Fact Verification module leverages the retrieved evidence and demonstrations to validate atomic facts, extracting key evidence to support the verification process and provide interpretable rationales for the results.

\subsection{Dynamic Atomic Fact Extraction}
The Dynamic Atomic Fact Extraction module serves as the foundational component of our fact verification framework, designed to decompose complex claims into simpler, atomic facts. 
This decomposition is critical for transforming intricate reasoning tasks into more manageable subtasks, enabling targeted and precise verification of individual facts. 
To address the inherent limitations of static decomposition strategies, we propose an iterative decomposition approach, which dynamically refines the extraction process based on previously verified facts, thereby mitigating the risk of error propagation and enhancing the overall reliability of the verification pipeline.

Given a complex claim \( C \), our approach incrementally extracts atomic facts, denoted as \( F = \{ F_1, F_2, \dots, F_T \} \), in a sequential manner. Instead of a static decomposition, where all atomic facts are generated all at once without incorporating verification feedback, our framework adopts an iterative extraction strategy that dynamically refines the process based on previously extracted and verified facts.

At each iteration \( t \), the model determines whether the extraction process should terminate or continue to extract the next atomic fact \( F_t \). This decision is based on assessing the coverage of the original claim \( C \) by the set of extracted atomic facts \( F_{1:t-1} \), ensuring that the decomposition remains both comprehensive and relevant. If the extracted facts collectively provide sufficient coverage of the claim, the process terminates. Otherwise, the system proceeds with extracting the next atomic fact \( F_t \).

When the extraction continues, \( F_t \) is generated by the LLM-based Extractor, which dynamically refines its output based on the understanding of previously extracted and verified facts. Specifically, the extractor generates predictions conditioned on the atomic facts from earlier iterations, denoted as \( F_{1:t-1} = \{F_1, F_2, \dots, F_{t-1}\} \), as well as their corresponding verification outcomes.
This iterative refinement process helps the model adapt its decomposition strategy, minimizing redundancy while ensuring that the extracted facts remain faithful to the claim structure. By incorporating verification results into the extraction loop, the framework reduces error propagation, improving both accuracy and robustness.

The fact extraction process can be formalized as a function where the extractor takes as input the original claim \( C \), along with the previously extracted facts and their verification results, to generate the next atomic fact:
\begin{equation}
F_t = \text{Extractor}(C, F_{1:t-1}, y_{1:t-1}, r_{1:t-1}).
\end{equation}

Here, \( F_{1:t-1} \) represents the set of atomic facts extracted up to iteration \( t \), while \( y_{1:t-1} \) and \( r_{1:t-1} \) captures their corresponding verification results and rationales. This formulation ensures that each newly extracted fact is informed by prior knowledge, facilitating a structured and adaptive claim decomposition process. Through this iterative methodology, the extraction process continuously refines its understanding of the claim while progressively verifying its individual components, leading to a more interpretable and reliable fact verification pipeline.

\subsection{Refined Evidence Retrieval}
To ensure reliable verification and mitigate hallucination risks, external evidence \( E_t \) is retrieved for each \( F_t \). It is obtained by Evidence Retrieval and Reranking, which is designed to ensure that the evidence used for verifying atomic facts is both highly relevant and free from noise. To further improve accurate reasoning, we incorporate Dynamic Instance Retrieval to identify related instances that align with each atomic fact \( F_t \). This allows the model to leverage highly relevant dynamic demonstrations and enhance the overall verification result.

\subsubsection{Evidence Retrieval and Reranking}

The retrieval process begins by identifying a preliminary set of candidate evidence \( E_t' \) for each atomic fact \( F_t \) from a large corpus \( D \). To capture semantic relevance, we compute the cosine similarity between \( F_t \) and each candidate \( e_j \in D \). The top-\( k' \) candidates with the highest similarity scores are selected as the initial retrieval set:
\begin{equation}
\text{score}(e_j, F_t) = \frac{f(e_j) \cdot f(F_t)}{\|f(e_j)\| \, \|f(F_t)\|},
\end{equation}
\begin{equation}
E_t' = \left\{ e_j \in D \,\middle|\, j \in \operatorname{arg\,top}_{k'}\, \text{score}(e_j, F_t) \right\},
\end{equation}
where \( f(\cdot) \) is a text encoder that produces dense vector embeddings. This retrieval step ensures coarse filtering of the corpus, yielding a candidate set for downstream reranking and reasoning.

However, due to the potential for irrelevant or noisy information in the initial retrieval stage, a pre-trained reranker is employed to filter and reorder the evidence. The reranker evaluates each piece of retrieved evidence and constructs a ranked subset of evidence \( E_t = \{e_t^1, e_t^2, \dots, e_t^k\} \) as follows:
\begin{equation}
E_t = \text{Reranker}(E_t', F_t).
\end{equation}

To optimize the reranker, we formulate a supervised training objective based on relevance judgments obtained from LLMs.
During training, instead of randomly sampling, we utilize LLMs to select positive and negative samples from the initial retrieved evidence set \( E_t' \) to enhance the model's discriminative ability.
Specifically, LLMs assess how each piece of evidence supports the verification of fact \( F_i \), assigning relevance scores accordingly.
The highest-scoring evidence is chosen as the positive sample \( e^+ \), and the \( m \) lowest-scoring pieces of evidence are selected as negative samples \( E^- = \{e_1^-, e_2^-, \dots, e_m^- \} \).
Subsequently, we apply an InfoNCE loss \( L_r\) as the learning objective of the reranker:
\begin{equation}
L_a = - \frac{1}{N} \sum_{i=1}^{N} \log \left( \frac{ \exp \left( f(F_i) \cdot f(e^+) \right) / \tau }{ \sum_{j=1}^{m} \exp \left( f(F_i) \cdot f(e_j^-) \right) / \tau } \right),
\end{equation}
where N is number of facts in each training batch, \( f(\cdot) \) is the encoder and \( \tau \) is the temperature coefficient.

This retrieval and reranking process ensures that only the most contributive evidence is retained, thereby reducing the impact of distractive or irrelevant information on the verification results.

\subsubsection{Dynamic Instance Retrieval}
In addition to retrieving evidence, the module dynamically identifies related claims and instances \( A_t \) that align with the atomic fact \( F_t \). These instances serve as dynamic demonstrations, providing in-context prompts that enhance the model's ability to reason about \( F_t \) by offering analogous examples and contextual guidance.
The Instance Retriever component searches for claims that share logical or semantic similarities with \( F_t \) from the training set, ensuring that the reasoning process is tailored to the unique characteristics of the fact under verification. This is achieved by computing the semantic similarity between \( F_t \) and the claims using a pre-trained language model, with claims that achieve high similarity scores being prioritized for retrieval. The retrieved instances \( A_t = \{ a_1^t, a_2^t, \dots, a_d^t \} \) are then used to guide the reasoning module, enabling it to generate more accurate and interpretable verification results.
The dynamic instance retrieval process can be formalized as follows:
\begin{equation}
\text{sim}(a_j, F_t) = \frac{f(a_j) \cdot f(F_t)}{\|f(a_j)\| \, \|f(F_t)\|},
\end{equation}
\begin{equation}
A_t = \left\{ a_j \in \mathcal{C} \mid j \in \operatorname{arg\,top}_d\, \text{sim}(a_j, F_t) \right\},
\end{equation}
where \( f(\cdot) \) is a text encoder and \( A_t \) represents the set of dynamically retrieved demonstrations for \( F_t \).

By incorporating these contextually relevant examples, the reasoning module is better equipped to handle complex and nuanced verification tasks, improving both the accuracy and interpretability of the final results. This approach not only enhances the model's reasoning capabilities but also ensures that the verification process is robust and adaptable to diverse claim structures and contexts.

\subsection{Adaptive Atomic Fact Verification}

The Adaptive Atomic Fact Verification Module constitutes the final and most critical stage of our fact verification framework, tasked with validating each atomic fact \( F_t \) using the retrieved external evidence \( E_t \) and dynamically retrieved instances \( A_t \).
The primary objective of this module is to generate a factuality label \( y_t \) (e.g., true, false, or unverifiable). 
To mitigate the potential mismatch in granularity between the fine-grained atomic fact \( F_t \) and the coarser instances, which may otherwise provide suboptimal contextual guidance, we incorporate the original claim \( C \) from which \( F_t \) is derived as an additional input to the verification module. 
This addition helps maintain a certain level of granularity alignment between the guidance and the fact.

In addition to the factuality label, the Reasoner produces a detailed rationale \( r_t \), which highlights the key pieces of evidence and reasoning steps that led to the conclusion. 
This rationale not only enhances the transparency and interpretability of the verification process but also provides valuable context for subsequent atomic fact extraction. 
For instance, the rationale may reveal implicit information (e.g., entity names or relationships) that refines the decomposition of the next atomic fact, leading to more precise and contextually grounded verification. 
This iterative feedback mechanism is crucial for handling complex claims, as it ensures that each step builds upon the insights gained from previous verifications.
The reasoning process can be formalized as follows:
\begin{equation}
y_t, r_t = \text{Reasoner}(F_t, C, E_t, A_t).
\end{equation}

By integrating evidence, demonstrations, and interpretable reasoning, this module not only delivers accurate verification outcomes but also fosters interpretability, making it a reliable system for handling complex claims.

Once all atomic facts \(F = \{F_1, F_2, \ldots, F_T \} \) have been verified, the final step is to aggregate the results to produce a comprehensive conclusion about the overall factuality of the original claim \( C \). 
The aggregation process combines the individual verification results \( y_t \) for each atomic fact \( F_t \) into a final claim validation \( y^* \). The aggregation process can be formalized as follows:
\begin{equation}
y^* = \text{Aggregate}\left( \{F_t\}_{t=1}^T, \{y_t\}_{t=1}^T, \{r_t\}_{t=1}^T \right),
\end{equation}
where \( y^* \) represents the final claim validation, and \( \{y_t\}_{t=1}^T \) denotes the set of factuality labels for all atomic facts.

Overall, by integrating external evidence \( E_t \), the Atomic Fact Verification module ensures that verification is anchored in reliable and relevant evidence, reducing the risk of hallucination. 
The use of dynamic demonstrations \( A_t \) further enhances the adaptability of the reasoning process, allowing the model to handle a wide range of claim types and complexities. 
Additionally, the generation of interpretable rationales \(r_t \) provides transparency, enhancing the transparency and interpretability of the verification process. 
Finally, the aggregation process ensures that the final claim validation \( y^* \) is both comprehensive and coherent, reflecting the overall truthfulness of the claim through its atomic verifications.

\subsection{Computational Complexity Analysis}

At the \( t \)-th iteration, we first invoke an LLM to generate the atomic fact \( F_t \), conditioned on the original claim, previously generated facts \( F_{1:t-1} \), rationales \( r_{1:t-1} \), labels \( y_{1:t-1} \), and a task-specific prompt. 
While invoking an LLM incurs non-negligible cost, the inherently compact nature of atomic facts and rationales enables us to control input size by limiting their lengths, thereby reducing the total number of tokens and improving inference efficiency.
Moreover, since no gradient computation is involved and each claim can be processed independently, this process is parallel and highly scalable during inference.
Similarly, in the verification stage for each atomic fact, we observe controllable inference costs and efficient LLM-driven reasoning, making the process both practical and scalable.

To improve efficiency in evidence retrieval, we adopt a two-stage approach. A bi-encoder retriever is first used to efficiently retrieve \( k' \) coarse candidate evidence snippets, with time complexity approximately \( O(\log N) \) for a corpus of size \( N \). These candidates are then reranked by a cross-encoder reranker with time complexity \( O(k'd) \), where \( d \) is the embedding dimension. Compared to using the reranker directly for precise retrieval, this cascaded strategy significantly reduces computational cost while maintaining retrieval quality.
Additionally, to reduce computational overhead during instance retrieval, we avoid searching over the full training set. Instead, we construct a candidate pool by randomly sampling a subset of training instances at a 1:1 ratio with the test set, preserving diversity while substantially improving retrieval efficiency.

In summary, our framework is designed with several efficiency- and cost-oriented strategies that enable scalable deployment while maintaining manageable inference overhead.

\section{Experiments}

\subsection{Experimental Setup} 
This section offers a detailed description of the datasets utilized for training and evaluation, the assessment metrics applied, the baseline models used for comparison, and the implementation details.

\subsubsection{Datasets.}
In this study, we leverage multiple publicly available fact verification datasets for model training and evaluation, including HOVER~\cite{HOVER}, PolitiHop~\cite{PolitiHop}, RAWFC~\cite{RAWFC}, LIAR~\cite{LIAR} and LIAR-PLUS~\cite{LIAR-PLUS}.
Specifically, the LIAR and LIAR-PLUS datasets are annotated with six distinct categories of truthfulness labels, enabling a fine-grained classification of claims based on their veracity. In contrast, HOVER, PolitiHop, and RAWFC are structured around three classification categories, providing a more general framework for fact verification tasks.

\begin{table}[ht]
\caption{Dataset Sizes for LIAR-PLUS, LIAR, HOVER, PolitiHop and RAWFC.}
\label{datasets}
    \centering
    \resizebox{0.45\columnwidth}{!}{
    \begin{tabular}{lccc}
        \toprule
         & Train & Dev & Test \\
        \midrule
        LIAR-PLUS & 10,240 & 1,284 & 1,267 \\
        LIAR & 10,269 & 1,284 & 1,283 \\
        HOVER & 18,171 & 4,000 & 4,000 \\
        PolitiHop & 592 & 141 & 200 \\
        RAWFC & 1,612 & 200 & 200 \\
        \bottomrule
    \end{tabular}
    }
    
\end{table}

\subsubsection{Baselines.}
To validate the effectiveness of the proposed AFEV model, we employ two distinct sets of existing models as baselines, ensuring a comprehensive evaluation. 
The first set of baselines comprises state-of-the-art multi-hop fact verification methods. 
As demonstrated in Table~\ref{multi-overall}, Pipeline~\cite{Pipeline} serves as a general-purpose baseline, offering a broad perspective on the task. IB~\cite{IB}, TSS~\cite{TSS}, and LR~\cite{LR} represent single-granular models, which focus exclusively on retrieving sentence-level evidence during the retrieval phase. 
Building upon insights from CURE~\cite{CURE}, we further enhance our baseline comparison by introducing a suite of multi-granular models. These models are designed to capture both sentence-level and token-level rationales, enabling a more nuanced and precise verification process. Specifically, DeClarE~\cite{DeClarE}, FRESH~\cite{FRESH}, SHAP~\cite{SHAP}, L-INTGRAD~\cite{L-INTGRAD}, DIFFMASK~\cite{DIFFMASK}, CURE~\cite{CURE}, and VMASK~\cite{VMASK} are integrated into our evaluation framework.  
This dual-level approach not only enhances the model's ability to discern subtle nuances in the evidence but also underscores the superiority of our proposed method in handling complex multi-hop reasoning tasks.

The second set of baselines consists of advanced LLM-based reasoning methods, which leverage the powerful capabilities of large language models for fact verification. As illustrated in Table~\ref{llm-overall}, these baselines include Standard Prompt~\cite{Standard-Prompt}, Vanilla CoT~\cite{Vanilla-CoT}, Search-Augmented CoT~\cite{HiSS}, ReAct~\cite{ReAct}, RAFTS~\cite{RAFV}, FFRR~\cite{FFRR} and HiSS~\cite{HiSS}. 
All of these methods rely heavily on LLMs but employ distinct reasoning strategies to achieve fact verification. Specifically, Standard Prompt directly queries the LLM to return the verification label for the claim.
In contrast, the other baselines adopt more sophisticated, step-by-step verification strategies, enabling the models to break down complex claims into intermediate reasoning steps.
To ensure a more granular and comprehensive comparison, we also incorporate several strong supervised models into our evaluation framework.
These models include CNN~\cite{CNN}, RNN~\cite{RNN}, DeClarE~\cite{DeClarE}, SentHAN~\cite{SentHAN}, SBERT~\cite{SBERT}, GenFE~\cite{GenFE}, and CofCED~\cite{CofCED}. 
These supervised models represent a diverse range of architectures and techniques, providing a robust benchmark for evaluating the performance of our proposed method. 


\subsubsection{Evaluation Metrics.}

For the multi-hop fact verification baselines, we conduct experiments on the LIAR-PLUS, HOVER, and PolitiHop datasets, ensuring alignment with established benchmarks in the field. To maintain consistency with prior works, particularly CURE~\cite{CURE}, we employ Label Accuracy and the F1 score as the primary evaluation metrics. These metrics are well-suited for assessing the model's ability to correctly classify claims while balancing precision and recall, which is critical for multi-hop reasoning tasks that require the integration of multiple evidence pieces.

For evaluating LLM-based reasoning methods, we utilize the LIAR and RAWFC datasets, which are widely recognized for evaluating the reasoning capabilities of language models. To ensure comparability with HiSS and RAFTS, we adopt precision, recall, and the F1 score as the evaluation metrics. These metrics provide a comprehensive assessment of the model's performance.

\subsubsection{Implementation Details}
Our framework employs GPT-3.5 to construct both the atomic fact extractor and the reasoner, leveraging its advanced natural language understanding and generation capabilities.
For the evidence retriever, we utilize a supervised fine-tuned language model, which is specifically trained to identify and retrieve relevant evidence from large corpora. To further enhance the precision of evidence retrieval, we use GPT-3.5 to sample both positive and negative examples, which are then used to train the evidence reranker. This approach ensures that the reranker can effectively distinguish between high-quality and low-quality evidence, improving the overall reliability of the verification process. Additionally, to ensure a fair and consistent comparison, all LLM-based baselines reported in Table~\ref{llm-overall} also use GPT-3.5 as their backbone. This alignment guarantees that the performance differences observed are solely attributable to the methodological advancements of our approach, rather than discrepancies in the underlying model capabilities.

\begin{figure*}[t]
\centering
\includegraphics[width=1.0\linewidth]{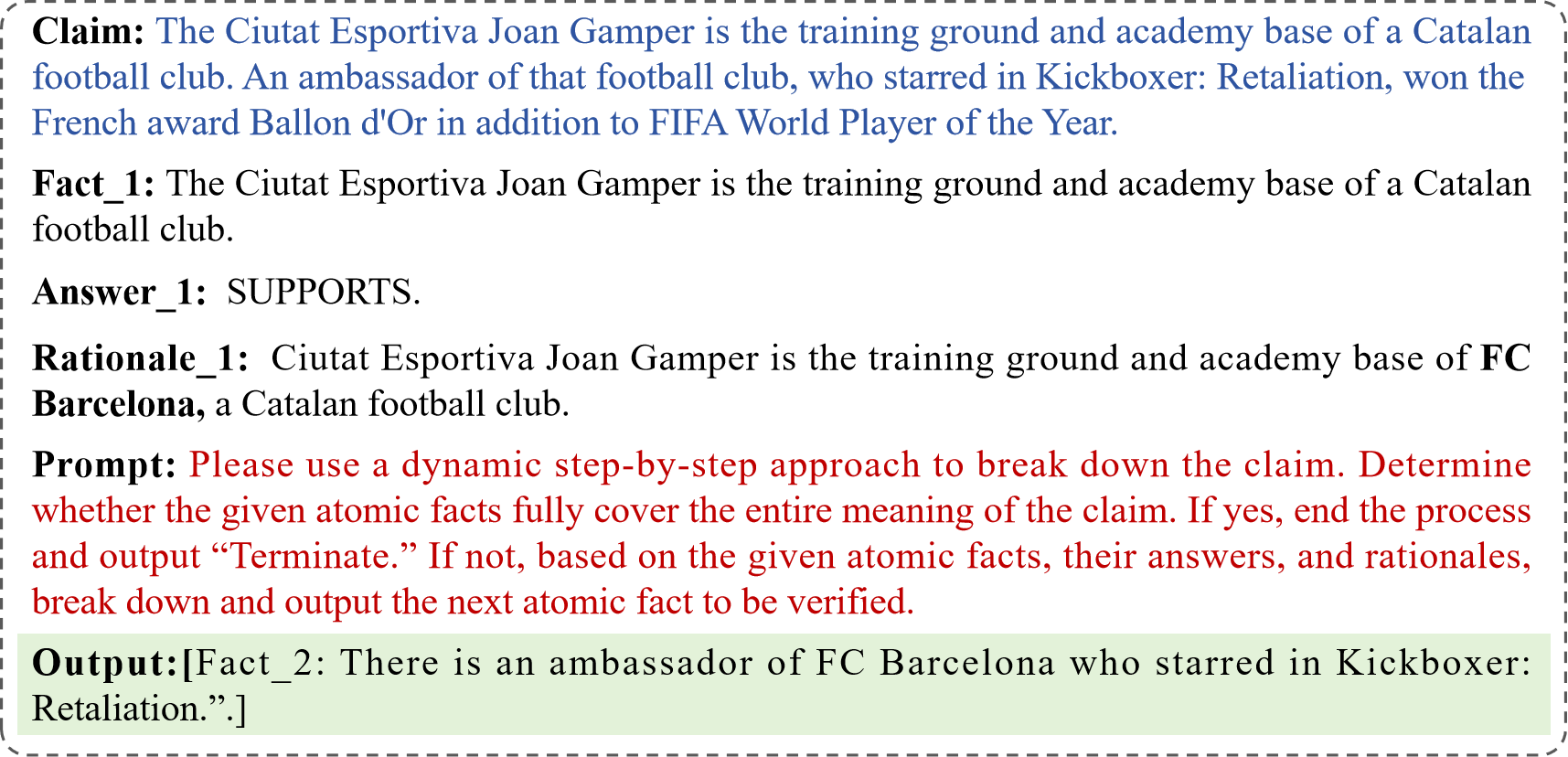}
\caption{Prompt for dynamic atomic fact extraction.}
\label{fig:prompt1}
\end{figure*} 

\begin{figure*}[t]
\centering
\includegraphics[width=1.0\linewidth]{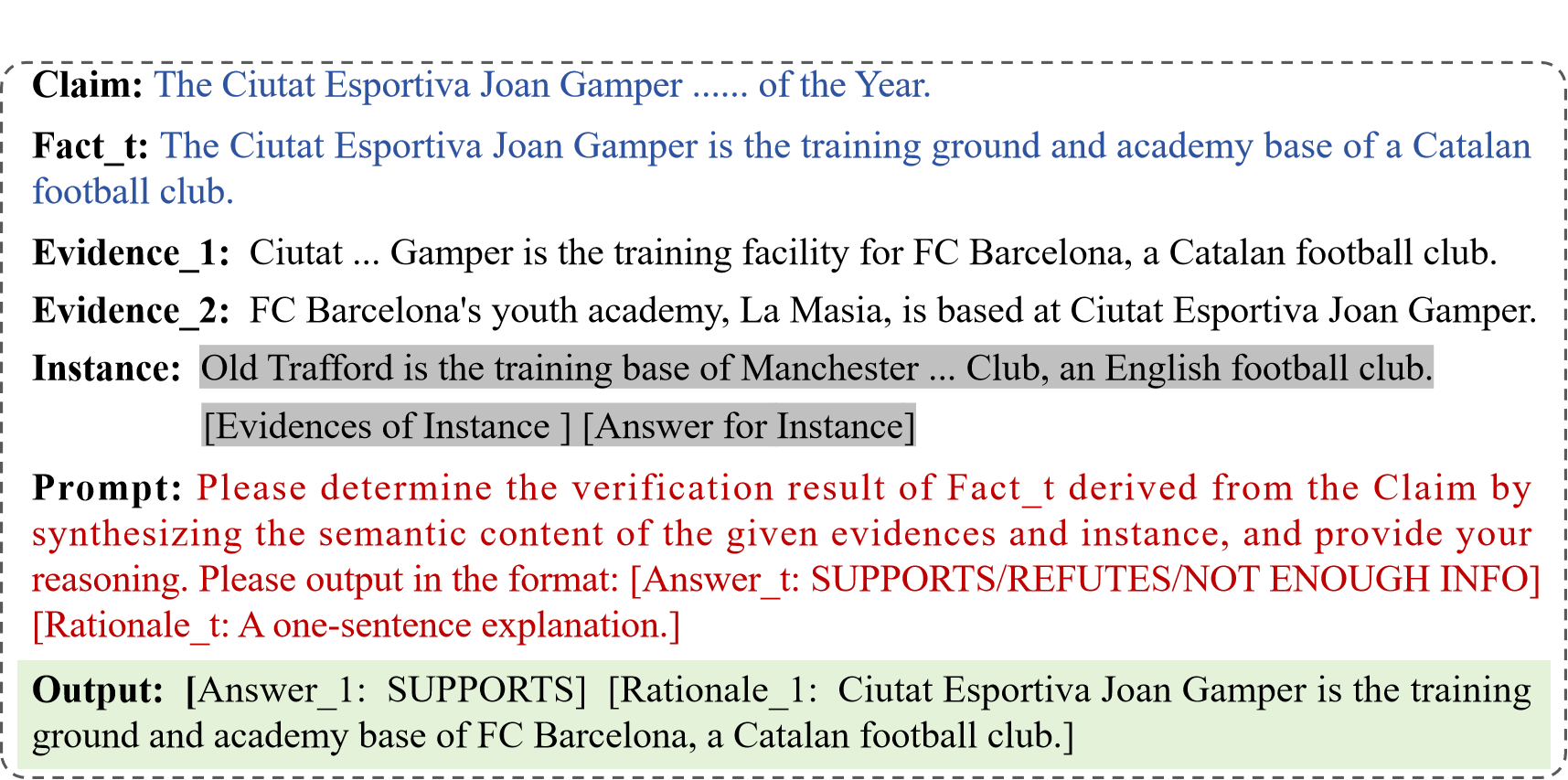}
\caption{Prompt for adaptive fact verification.}
\label{fig:prompt2}
\end{figure*}

After extensive parameter tuning, we set the final number of evidence retrieval candidates to 5 and the number of reranked evidence to 2. Additionally, for each atomic fact to be verified, we dynamically select 1 demonstration to guide the reasoning process. This dynamic demonstration selection ensures that the model adapts to the specific context of each claim, enhancing its ability to handle diverse and complex verification scenarios.
Moreover, Figure~\ref{fig:prompt1} and Figure~\ref{fig:prompt2} provide illustrative examples of the prompts employed for dynamic atomic fact extraction and adaptive fact verification, respectively.

\subsection{Overall Performance}

Compared to single-granular models, multi-granular methods, particularly CURE and VMASK, demonstrate the ability to capture fine-grained evidence. 
These methods not only aggregate a richer set of clues by leveraging both sentence-level and token-level rationales, but also reduce the reasoning burden on the model through the explicit use of token-level evidence, which provides more precise and localized information. 
As a result, as illustrated in Table~\ref{multi-overall}, multi-granular methods generally outperform traditional single-granular models in fact verification tasks, achieving higher accuracy.
However, our proposed AFEV method, despite relying on sentence-level single-granular evidence for each atomic fact verification, achieves state-of-the-art performance. 
This remarkable result highlights that while fine-grained evidence can enhance performance, the design of the retrieval and reasoning mechanisms also plays a critical role in achieving superior results.

\begin{table*}[ht]
\caption{Overall performance on LIAR-PLUS, HOVER and PolitiHop.
The improvement over the best baseline method (i.e., VMASK) is statistically significant (sign test, p-value $<$ 0.01). The best results are highlighted in \textbf{bold}, and the second-best results are \underline{underlined}.}
\label{multi-overall}
\begin{center}
\setlength{\tabcolsep}{5mm}{
\begin{normalsize}
\renewcommand{\arraystretch}{1.2}
\resizebox{1.00\columnwidth}{!}{
\begin{tabular}{lcccccc}
\toprule
         & \multicolumn{2}{c}{\textbf{LIAR-PLUS}} & \multicolumn{2}{c}{\textbf{HOVER}} & \multicolumn{2}{c}{\textbf{PolitiHop}}\\ \cmidrule(r){2-3} \cmidrule(r){4-5}  \cmidrule(r){6-7}
\textbf{Models} & \textbf{LA}    & \textbf{Macro-F1}    & \textbf{LA}      & \textbf{Macro-F1}    & \textbf{LA}     & \textbf{Macro-F1}   \\ 

\midrule
Pipeline  & 58.11 & 53.93  & 62.55 & 62.44   & 65.96 & 41.73\\
IB       & 62.52 & 60.48 & 56.78 & 56.74    & 68.79 & 54.89\\
TSS      & 62.39 & 61.72  & 53.68 & 51.11   & 65.25 & 43.34\\
LR       & 76.52 & 75.19  & 51.10 & 40.50   & 70.21 & 47.12\\

\midrule
DeClarE  & 47.73    & 21.54  & 50.83       & 50.76  & 69.50   & 27.34\\
FRESH  & 43.45    & 41.37  & 60.28 &  60.14  & 61.70    & 44.35\\

SHAP  & 76.39   & 75.31 &  59.83   &  58.18  & 59.57    & 40.71\\
L-INTGRAD  & 71.72   &  69.84  & 50.03   & 53.86  & 69.50    & 27.34\\
DIFFMASK  & 58.50   &  48.03 &  71.53     & 71.30 & 67.38    & 44.71\\
CURE  & 82.10    &  80.78  & \underline{76.98}   & \underline{76.89}   & 69.50    & 32.36\\
VMASK  & \underline{82.62}  & \underline{81.46}  & 74.38    & 73.69   & \underline{72.34}    & \underline{55.80}\\

 \midrule
 \textbf{AFEV}  & \textbf{83.73}          & \textbf{83.12}  & \textbf{78.87}    & \textbf{78.76} & \textbf{74.14}  & \textbf{57.69}   \\

\bottomrule
\end{tabular}}
\end{normalsize}
}
\end{center}
\end{table*}

Since AFEV utilizes LLMs for atomic fact extraction and verification, we further compare it with LLM-based step-by-step baselines to highlight  its superiority. 
We choose to evaluate on LIAR and RAWFC because they represent particularly challenging verification scenarios where traditional supervised pipelines often struggle.
LIAR lacks annotated evidence, and RAWFC provides only weakly structured evidence, making both datasets better suited for assessing the effectiveness of LLM-based approaches.
As shown in Table~\ref{llm-overall}, despite the remarkable reasoning capabilities of LLMs, models relying on simple reasoning strategies still underperform compared to supervised fact verification models. 
This performance gap may stem from limitations in evidence retrieval quality and the inherent hallucination issues of LLMs, which can lead to unreliable or factually incorrect outputs. 
To address these challenges, methods like RAFTS and HiSS have introduced enhanced retrieval and reasoning strategies, guiding LLMs to achieve improved performance in fact verification tasks. 
However, our proposed model surpasses even these advanced baselines, achieving state-of-the-art results on both the RAWFC and LIAR datasets. 
This superior performance demonstrates the effectiveness of our approach in integrating evidence retrieval, reranking, and reasoning while effectively addressing the limitations of LLMs.

\begin{table*}[ht]
\caption{Overall performance on LIAR and RAWFC.
The improvement over the best baseline method (i.e., RAFTS) is statistically significant (sign test, p-value $<$ 0.01).}
\label{llm-overall}
\begin{center}
\setlength{\tabcolsep}{5mm}{
\begin{normalsize}
\renewcommand{\arraystretch}{1.2}
\resizebox{0.9\columnwidth}{!}{
\begin{tabular}{lcccccc}
\toprule
         & \multicolumn{3}{c}{\textbf{RAWFC}} & \multicolumn{3}{c}{\textbf{LIAR}}\\ \cmidrule(r){2-4} \cmidrule(r){5-7}  
\textbf{Models} & \textbf{P}    & \textbf{R}    & \textbf{F1}      & \textbf{P}    & \textbf{R}     & \textbf{F1}   \\ 

\midrule
CNN & 38.8 & 38.5 & 38.6 & 22.6 & 22.4 & 22.5 \\
RNN  & 41.4 & 42.1 & 41.7 & 24.4 & 21.2 & 22.7 \\
DeClarE & 43.4 & 43.5 & 43.4 & 22.9 & 20.6 & 21.7 \\
SentHAN & 45.7 & 45.5 & 45.6 & 22.6 & 20.0 & 21.2 \\
SBERT & 51.1 & 46.0 & 48.4 & 24.1 & 22.1 & 23.1 \\
GenFE & 44.3 & 44.8 & 44.5 & 28.0 & 26.2 & 27.1 \\
CofCED & 53.0 & 51.0 & 52.0 & 29.5 & 29.6 & 29.5 \\

\midrule
Standard Prompt & 48.5 & 48.5 & 48.5 & 29.1 & 25.1 & 27.0 \\
Vanilla CoT & 42.4 & 46.6 & 44.4 & 22.6 & 24.2 & 23.7 \\
Search-Augmented CoT & 47.2 & 51.4 & 49.2 & 27.5 & 23.6 & 25.4 \\
ReAct & 51.2 & 48.5 & 49.8 & 33.2 & 29.0 & 31.0 \\
HiSS & 53.4 & 54.4 & 53.9 & 46.8 & 31.3 & 37.5 \\
FFRR & 56.5 & \underline{57.4} & 57.0 & 34.5 & 32.6 & 33.5 \\
RAFTS & \underline{62.8} & 52.6 & \underline{57.3} & \underline{47.1} & \underline{37.9} & \underline{42.0} \\


\midrule
 \textbf{AFEV}  & \textbf{63.3}          & \textbf{57.6}  & \textbf{60.2}    & \textbf{48.2} & \textbf{40.3}  & \textbf{43.9}   \\

\bottomrule
\end{tabular}}
\end{normalsize}
}
\end{center}
\end{table*}

\subsection{Performance on Evidence Retrieval}

\begin{table}[ht]
\caption{Retrieval performance comparison.}
\label{evi-hover}
\begin{center}
\begin{small}
\resizebox{0.6\columnwidth}{!}{
\begin{tabular}{llcccc}
\toprule
\textbf{ } & \textbf{Models} & \textbf{F1}          & \textbf{Precision}        & \textbf{Recall}  \\ 

\midrule
\multirow{6}{*}{\textbf{LIAR-PLUS}} 

& Pipeline  & 0.6677 & 0.7450 & 0.6564 \\ 
& IB   & 0.3777 & 0.3927 & 0.3967 \\ 
& TSS  & 0.4324 & 0.6349 & 0.3469 \\ 
& LR   & 0.6242 & 0.6776 & 0.6381 \\ 
& CURE & 0.6789 & 0.8072 & 0.6329 \\
& \textbf{AFEV}   & \textbf{0.6952}  & \textbf{0.8342}  &  \textbf{0.6427}\\

\midrule
\multirow{6}{*}{\textbf{HOVER}} 
& Pipeline & 0.9427 & 0.9028 & 0.9900 \\ 
& IB    & 0.6236 & 0.7018 & 0.5783 \\ 
& TSS    & 0.6883 & 0.9026 & 0.5755 \\ 
& LR    & 0.9419 & 0.9029 & \textbf{0.9988} \\ 
& CURE & 0.9376 & 0.9045 & 0.9877 \\ 
& \textbf{AFEV}   & \textbf{0.9423}  & \textbf{0.9096}  &  \textbf{0.9923}\\

\midrule
\multirow{6}{*}{\textbf{PolitiHop}} 
& Pipeline  & 0.6390 & 0.5986 & 0.8234 \\ 
& IB   & 0.4180 & 0.5106 & 0.3902 \\ 
& TSS    & 0.4272 & 0.5177 & 0.4044 \\ 
& LR    & 0.5699 & 0.5674 & 0.6657 \\ 
& CURE & 0.6947 & 0.6584 & 0.8403 \\ 
& \textbf{AFEV}   & \textbf{0.7021}  & \textbf{0.6737}  &  \textbf{0.8137}\\

\bottomrule
\end{tabular}}
\end{small}
\end{center}
\end{table}

Evidence retrieval is an indispensable component of fact verification tasks, as it directly impacts the quality and reliability of the verification process. Since AFEV retrieves sentence-level evidence, we aim to validate the effectiveness of our retrieve-then-rerank strategy by comparing its performance with single-granular models, following the methodology of prior work~\cite{CURE}. Specifically, we conduct experiments on the HOVER, LIAR-PLUS and PolitiHop dataset, which is well-suited for evaluating multi-hop retrieval capabilities.
As shown in Table~\ref{evi-hover}, AFEV achieves state-of-the-art retrieval performance across all datasets, outperforming existing single-granular baselines. This superior performance underscores the effectiveness of our retrieve-then-rerank strategy, which combines comprehensive retrieval with precise reranking to identify the most relevant and high-quality evidence.

\subsection{Ablation Study}

To thoroughly evaluate the contribution of each component in AFEV, we conduct a series of ablation experiments on the HOVER dataset. 
These experiments are designed to isolate and analyze the impact of key modules within our framework, providing insights into their individual and collective roles in achieving superior performance.
The ablation settings are described as follows:

\textbf{-w/o Atomic Fact Extraction:}
We skip the decomposition of initial complex claims and directly verify the claims without breaking them down into atomic facts. 
This ablation is designed to demonstrate the necessity of atomic fact extraction in handling complex claims, particularly in multi-hop reasoning scenarios where decomposing claims into simpler units is essential for accurate verification.

\textbf{-w/o Iterative Extraction:}
The iterative decomposition strategy is abandoned, and the initial complex claims are decomposed into atomic facts in a single step. 
This setting aims to highlight the importance of iterative refinement in accurately identifying and extracting atomic facts, as a one-shot decomposition may overlook nuanced dependencies and relationships within the claim.

\textbf{-w/o Rationales:}
The use of rationales generated during intermediate steps of the iterative decomposition process is removed. Specifically, the model does not leverage  the explanatory information about verification results from previous steps to guide the extraction of subsequent atomic facts. 
This ablation is intended to evaluate the role of explanatory feedback in improving the precision and coherence of atomic fact extraction.


\textbf{w/o Reranking:} 
We disable the fine-tuned reranker and directly use the initially retrieved evidence without further filtering. This setting aims to highlight the importance of reranking in reducing noise and improving the quality of evidence, as unfiltered evidence may contain irrelevant or low-quality content that hinders accurate verification.

\textbf{w/o Dynamic Demonstrations:} 
The dynamic demonstrations are replaced with static ones, which remain fixed across different atomic facts. This ablation is designed to highlight the benefits of dynamic demonstration selection, which adapts to the specific context of each atomic fact, enhancing the relevance and effectiveness of the reasoning process.

\textbf{w/o Evidence Retrieval:} 
We disable evidence retrieval from the external corpus but still sample dynamic demonstrations to guide the verification process. This ablation aims to isolate the contribution of external evidence retrieval, demonstrating how the model performs when relying solely on demonstrations and internal reasoning capabilities.

\textbf{w/o Demonstrations:} 
This variant disables the similar instances used for in-context demonstrations.
This setting evaluates the role of demonstration-based reasoning in providing contextual guidance and improving the model's ability to handle complex claims by leveraging analogous examples.

\textbf{w/o External Evidence Retrieval:} 
We completely remove both evidence retrieval and demonstration sampling, relying exclusively on the reasoner's internal parametric knowledge (e.g., GPT-3.5) to verify each atomic fact. This ablation evaluates the model's ability to perform fact verification without any external evidence or contextual guidance, highlighting the limitations of relying solely on the LLMs' inherent knowledge.

\begin{table}[ht]
\caption{Performance of ablation study on HOVER.}
\label{ablation}
\begin{center}
\begin{small}
\resizebox{0.85\columnwidth}{!}{
\begin{tabular}{clcc}
\toprule
\textbf{ } & \textbf{Models} & \textbf{LA}   & \textbf{Macro-F1}   \\ 


\midrule
\multirow{3}{*}{\textbf{\makecell{Atomic Fact \\ Extraction}}} 
& -w/o Atomic Fact Extraction  & 77.04 & 76.96 \\
& -w/o Iterative Extraction  &  77.66 & 77.81  \\
& -w/o Rationales  & 78.74  & 78.82 \\

\midrule
\multirow{2}{*}{\textbf{\makecell{External Evidence \\ Retrieval}}} 
& -w/o Reranking  & 77.97 & 78.04 \\
& -w/o Dynamic Demonstrations & 78.35 &  78.51 \\

\midrule
\multirow{3}{*}{\textbf{\makecell{Atomic Fact \\ Verification}}} 
& -w/o Evidence Retrieval   & 77.35  & 77.29  \\
& -w/o Demonstrations  & 78.20 & 78.36 \\
& -w/o External Evidence Retrieval & 77.16 & 77.09  \\

\midrule
\multirow{1}{*}{ } 
& \textbf{AFEV}  &\textbf{78.87}        & \textbf{78.76}         \\


\bottomrule
\end{tabular}}
\end{small}
\end{center}
\end{table}

As shown in Table~\ref{ablation}, the removal or modification of each component leads to a noticeable decline in overall model performance, underscoring their importance within the overall framework.
Specifically, the result of -w/o Atomic Fact Extraction demonstrates that decomposing complex claims into atomic facts significantly enhances model performance for fact verification. 
It not only reduces the reasoning complexity by breaking down complex claims into manageable units but also facilitates the retrieval of more comprehensive and relevant external evidence.
In addition, the proposed iterative extraction strategy further improves performance by mitigating error accumulation throughout the decomposition process, and this incremental refinement is critical for handling complex claims with multiple layers of reasoning.

Retrieving evidence from the external knowledge corpus is essential for mitigating hallucination issues inherent in LLMs, as it provides factual grounding and reduces reliance on the model's parametric knowledge alone. Therefore, removing evidence retrieval in -w/o Evidence Retrieval results in a significant decrease in verification performance. Additionally, the results of w/o Demonstrations and -w/o Dynamic Demonstrations emphasize the importance of context-aware demonstration selection. Dynamic demonstrations are further adapt to the specific context of each atomic fact, ensuring that the reasoning process is guided by the most relevant and informative examples. This adaptability enhances the model's ability to handle diverse and nuanced verification scenarios.

\subsection{Hyperparameter Sensitivity Analysis}

We conduct a hyperparameter sensitivity analysis on the HOVER dataset to evaluate the impact of two critical parameters: \( k \), the number of evidence pieces selected after evidence reranking, and \( d \), the number of dynamically sampled similar instances used as reasoning demonstrations.

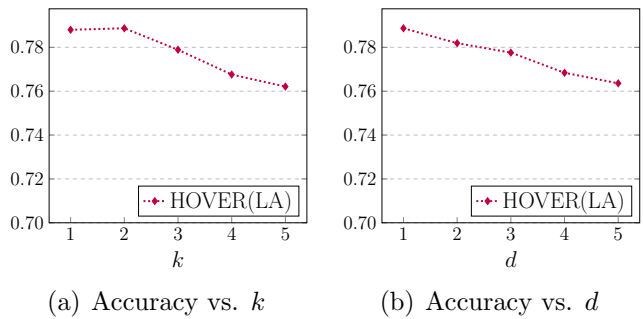
\begin{figure}[ht]
\centering
    \subfigure[Accuracy vs. $k$]{\label{parameters-k}
        \begin{tikzpicture}[font=\Large,scale=0.5]
            \begin{axis}[
                legend cell align={left},
                legend style={nodes={scale=1.0, transform shape}},
                xlabel={$k$},
                xtick pos=left,
                tick label style={font=\large},
                ylabel style={font=\large},
                ylabel={ },
                ymin=0.70,
                xtick={1, 2, 3, 4, 5},
                xticklabels={$1$, $2$, $3$, $4$, $5$},
                legend pos=south east,
                ymajorgrids=true,
                grid style=dashed,
                y tick label style={/pgf/number format/precision=2, /pgf/number format/fixed, /pgf/number format/fixed zerofill}
            ]
            \addplot[
                color=purple,
                dotted,
                mark options={solid},
                mark=diamond*,
                line width=1.5pt,
                mark size=2pt
                ]
                coordinates {
                (1, 0.7880)
                (2, 0.7887)
                (3, 0.7789)
                (4, 0.7676)
                (5, 0.7621)
                };
            \addlegendentry{HOVER(LA)}
            \end{axis}
        \end{tikzpicture}
    }
    \subfigure[Accuracy vs. $d$]{\label{parameters-d}
        \begin{tikzpicture}[font=\Large,scale=0.5]
            \begin{axis}[
                legend cell align={left},
                legend style={nodes={scale=1.0, transform shape}},
                xlabel={$d$},
                xtick pos=left,
                tick label style={font=\large},
                ylabel style={font=\large},
                ylabel={ },
                ymin=0.70,
                xtick={1, 2, 3, 4, 5},
                xticklabels={$1$, $2$, $3$, $4$, $5$},
                legend pos=south east,
                ymajorgrids=true,
                grid style=dashed,
                y tick label style={/pgf/number format/precision=2, /pgf/number format/fixed, /pgf/number format/fixed zerofill}
            ]
            \addplot[
                color=purple,
                dotted,
                mark options={solid},
                mark=diamond*,
                line width=1.5pt,
                mark size=2pt
                ]
                coordinates {
            (1, 0.7887)
            (2, 0.7819)
            (3, 0.7776)
            (4, 0.7684)
            (5, 0.7636)
                };
            \addlegendentry{HOVER(LA)}
            \end{axis}
        \end{tikzpicture}
    }
    \caption{Hyperparameter sensitivity analysis.}
\label{fig:para}
\end{figure}

As depicted in Figure~\ref{parameters-k}, the results indicate that when \( k \) is too small, the model suffers from insufficient evidence, leading to incomplete or inaccurate verification results. 
Conversely, when \( k \) is too large, the model is exposed to excessive noise, as irrelevant or low-quality evidence degrades performance.
These findings suggest that selecting 1–2 pieces of evidence per atomic fact strikes the best balance, as decomposing complex claims into atomic facts reduces the amount of information to verify, requiring fewer yet more precise evidence pieces.

Similarly, the number of demonstrations \( d \) plays a crucial role in guiding the reasoning process. Using too few demonstrations (e.g., \( d = 1 \)) often results in insufficient contextual guidance, while too many (e.g., \( d > 2 \)) introduce redundancy and unnecessary complexity. Results in Figure~\ref{parameters-d} show that 1-2 demonstrations per atomic fact are sufficient for reasoning performance. This is attributed to the reduced reasoning complexity of atomic facts, which requires less contextual support.

\subsection{Efficiency Study}

\begin{table}[ht]
\caption{Performance of efficiency study on HOVER.}
\label{efficiency}
\begin{center}
\begin{small}
\resizebox{0.85\columnwidth}{!}{
\begin{tabular}{clccc}
\toprule
\textbf{ } & \textbf{Models} & \textbf{LA}   & \textbf{Macro-F1}  & \textbf{Time/(h)} \\

\midrule
\multirow{2}{*}{\textbf{\makecell{One-hop \\ Extraction}}} 
& -w/o Iterative Extraction  & 77.04 & 76.96  & 0.86 \\  
& -w/o Iterative Extraction and  Reranking  &  76.84 & 76.86 & 0.71  \\

\midrule
\multirow{2}{*}{\textbf{\makecell{Iterative \\ Extraction}}} 
&  -w/o Reranking  & 77.97 & 78.04  & 0.79 \\
& \textbf{AFEV}  &\textbf{78.87}        & \textbf{78.76}   &  0.94\\

\bottomrule
\end{tabular}}
\end{small}
\end{center}
\end{table}

To assess both the computational efficiency of our model, we evaluate AFEV and its ablations on the HOVER test set. As summarized in Table~\ref{efficiency}, \textit{-w/o Iterative Extraction} replaces the iterative decomposition strategy with a one-shot approach. While this variant slightly improves runtime, it results in a noticeable decline in verification accuracy.
In contrast, \textit{-w/o Reranking} retains iterative decomposition but omits the evidence reranking stage. Despite relying on less precise evidence retrieval, it outperforms \textit{-w/o Iterative Extraction}, highlighting the value of accurate atomic claim generation through iterative decomposition.
Besides, \textit{-w/o Reranking} also achieves lower runtime than \textit{-w/o Iterative Extraction}, indicating that iterative decomposition does not substantially compromise efficiency. This alleviates concerns that its iterative nature would significantly hinder overall runtime performance.

\subsection{Case Study}

\begin{figure*}[ht]
\centering
\includegraphics[width=1.0\linewidth]{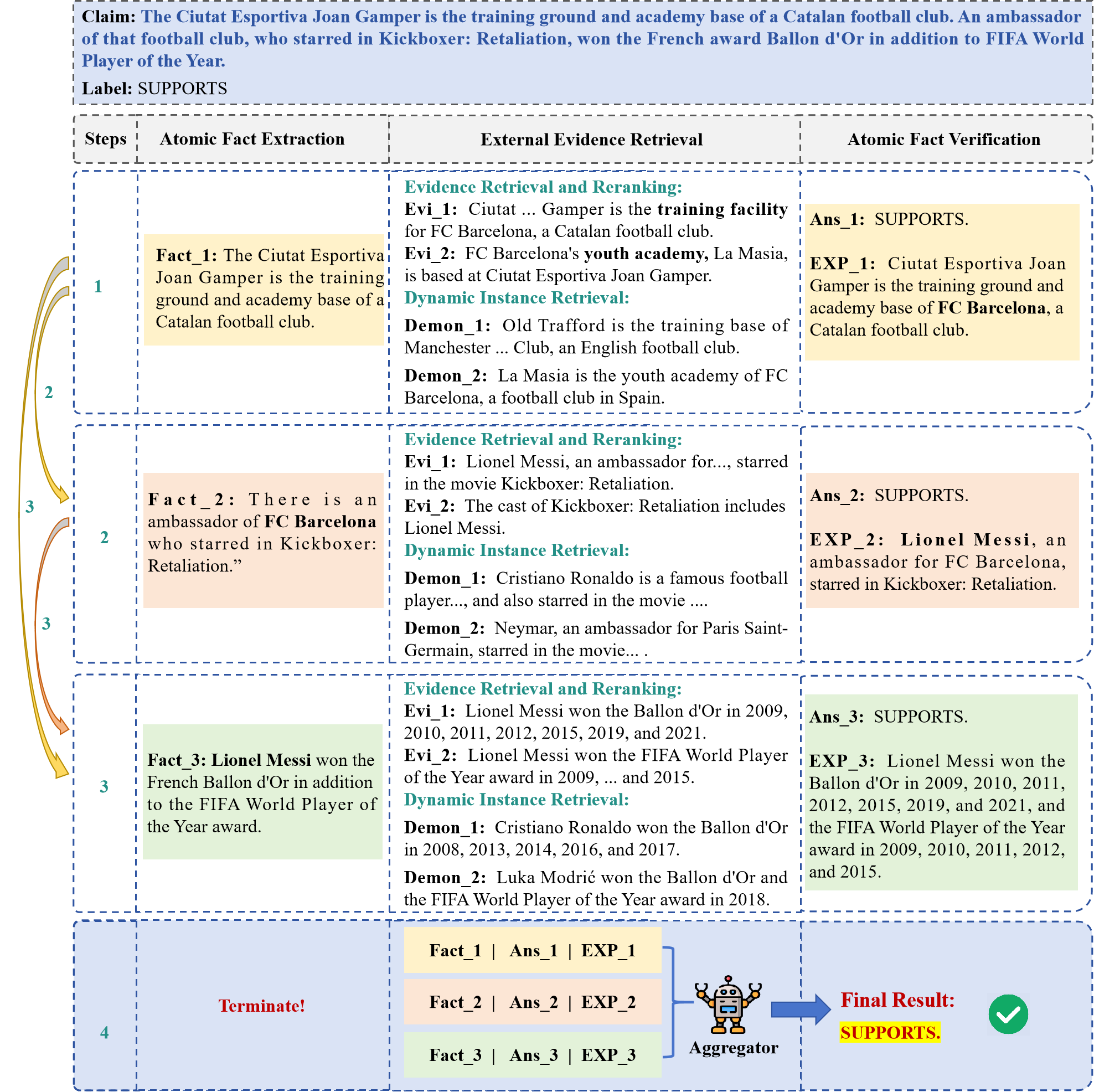}
\caption{The reasoning process of AFEV for a specific case.}
\label{fig:casestudy}
\end{figure*}  

In this section, we present a case study to systematically illustrate the execution mechanism of the proposed AFEV model. As depicted in Figure~\ref{fig:casestudy}, the given claim contains multiple pieces of information, making it challenging to verify in a single step. 
Under our framework, AFEV first extracts the initial atomic fact (Fact1) from the original claim and retrieves key external evidence to support its verification, along with dynamic demonstrations to guide the reasoning process. 
Subsequently, leveraging the verification result and rationale of Fact1, the model extracts the second atomic fact (Fact2). 
Unlike direct decomposition, this iterative strategy refines Fact2 by replacing “that football club” with a more specific entity, “FC Barcelona.” 
This improves clarity and facilitates more accurate evidence retrieval and reasoning.
Following a similar verification process as Fact1, the model easily obtains the verification result and rationale for Fact2.

Building on the enhanced context from Fact1 and Fact2, the extracted Fact3 further incorporates additional information not explicitly present in the original claim, such as “Lionel Messi,” which is derived from the rationale of Fact2. 
This demonstrates the high rationality and superiority of the proposed iterative extraction and reasoning strategy. 
Finally, once all sub-information in the original claim has been fully extracted and verified, the extractor automatically terminates the decomposition process and synthesizes the verification results of all atomic facts to produce the final claim verification result.

This case study highlights the effectiveness of our framework in handling complex claims through iterative atomic fact extraction, evidence retrieval, and dynamic demonstration-guided reasoning, ensuring both accurate and interpretable verification results.

\section{Conclusion}
This paper presents AFEV, a novel framework for fact verification that addresses the challenges of complex claims through dynamic atomic fact extraction, refined evidence retrieval and adaptive atomic fact verification. 
By decomposing intricate claims into simpler atomic facts and leveraging external evidence alongside contextually relevant demonstrations, AFEV significantly enhances the accuracy and interpretability of the verification process. 
The integration of iterative extraction and adaptive reasoning mechanisms ensures that each subcomponent of the claim is meticulously verified and contextualized, ultimately leading to a more reliable overall result.
Extensive experiments on five benchmark datasets demonstrate that AFEV achieves state-of-the-art performance, underscoring its effectiveness in handling nuanced verification tasks.

\section{Acknowledgements}
This work was supported by the Natural Science Foundation of China [No. 62372057].











\end{document}